\documentclass[letterpaper, 10 pt, conference]{ieeeconf}  
\pdfoutput=1

\IEEEoverridecommandlockouts                              

\usepackage{aecompl}
\usepackage{subfigure}
\usepackage{booktabs} 
\usepackage{graphics} 
\usepackage{epsfig} 
\usepackage{mathptmx} 
\usepackage{amsmath} 
\usepackage{amssymb}  
\usepackage{multirow} 
\usepackage{mathrsfs}
\usepackage{amsfonts}
\usepackage{enumerate}
\usepackage{mathtools,amssymb}
\usepackage{makeidx}
\usepackage{ntheorem}
\usepackage{cite}
\usepackage{url}
\theorembodyfont{\rmfamily}
\theoremseparator{:}

\makeindex

\title{\LARGE \bf
 Learning-Based Safety-Stability-Driven Control for Safety-Critical Systems under Model Uncertainties
}

\author{Lei Zheng, Rui Yang, Jiesen Pan, Hui Cheng$^{*}$, and Haifeng Hu
	\thanks{L. Zheng and H. Hu are with the School of Electronics and Information Technology, Sun Yat-sen University, Guangzhou 510006, China.}%
	\thanks{R. Yang, J. Pan, H. Cheng are with the School of Data and Computer Science, Sun Yat-sen University, Guangzhou 510006, China.}%
	\thanks{*Corresponding author: chengh9@mail.sysu.edu.cn}%
}
\begin{document}
	
	\maketitle
	\thispagestyle{empty}
	\pagestyle{empty}

	\begin{abstract}
Safety and tracking stability are crucial for safety-critical systems such as self-driving cars, autonomous mobile robots, and industrial manipulators. To efficiently control safety-critical systems to ensure their safety and achieve tracking stability, accurate system dynamic models are usually required. However, accurate system models are not always available in practice. In this paper, a learning-based safety-stability-driven control (LBSC) algorithm is presented to guarantee the safety and tracking stability for nonlinear safety-critical systems subject to control input constraints under model uncertainties. Gaussian Processes (GPs) are employed to learn the model error between the nominal model and the actual system dynamics, and the estimated mean and variance of the model error are used to quantify a high-confidence uncertainty bound. Using this estimated uncertainty bound, a safety barrier constraint is devised to ensure safety, and a stability constraint is developed to achieve rapid and accurate tracking. Then the proposed LBSC method is formulated as a quadratic program incorporating the safety barrier, the stability constraint, and the control constraints. The effectiveness of the LBSC method is illustrated on the safety-critical connected cruise control (CCC) system simulator under model uncertainties. 

	\end{abstract}
	
	\section{Introduction}
	Safety is a fundamental issue for safety-critical systems~\cite{hovakimyan20111}, such as self-driving cars, autonomous mobile robots, industrial manipulators, chemical reactors. Safety-critical systems may operate in unsafe states due to model inaccuracies, external disturbances, and environmental changes. It will lead to unexpected failures and severe damage to machines, environments, and even human life~\cite{Knight2002SafetyCS}. 
	
	On the other hand, achieving stable tracking performance with respect to tracking accuracy and convergence rate is crucial for dynamical control systems to accurately follow the desired states. However, there is a tradeoff between safety and tracking performance, and it is difficult for safety-critical systems with control input constraints to simultaneously satisfy safety and stabilization constraints in practical applications. For example, it is challenging for a self-driving car to rapidly converge to the desired speed and maintain it while avoiding unexpected obstacles. For a team of safety-critical vehicles, each vehicle keeps tracking its front vehicle at a desired constant speed while maintaining a safe following distance with it in normal situations. However, when a vehicle decelerates urgently in unexpected situations, the vehicle behind has to violate its stability constraint and reduce its speed to avoid collision with the front vehicle. In these cases, there exists a conflict between safety and stable high-performance tracking. For safety-critical systems, safety must not be violated and the tracking errors should be kept as small as possible. Hence, it is important to develop efficient methods for safety-critical systems to mediate the tradeoff between safety and tracking performance.
	
	Accurate system models are usually required to achieve safety and accurate tracking for dynamical control systems. However, exact system models are hard to obtain or even unavailable in practical applications. To address this challenging problem, Gaussian Processes (GPs) have been incorporated into Model Predictive Control (MPC) to account for model uncertainties~\cite{Teck2018GaussianPA, Ostafew2016RobustCL}. However, it is nontrivial to specify a proper cost function~\cite{schaal2010learning} to handle safety and tracking performance tradeoff.
	
	Alternatively, the CBF-CLF-QPs approaches~\cite{Ames2019ControlBF} incorporating control barrier functions (CBFs) and control Lyapunov functions (CLFs) based on quadratic programs (QPs) are promising to mediate the tradeoff between safety and tracking stability, with the safety being guaranteed. The CBF-CLF-QPs framework has achieved great success in varieties of safety-critical systems such as adaptive cruise control systems~\cite{Ames2017ControlBF,Xu2015RobustnessOC}, freedom bipedal robots~\cite{Agrawal2017DiscreteCB}, and quadrotor systems~\cite{wu2016safety}. Nevertheless, in the presence of model uncertainties, the safety constraints and tracking stability constraints may be violated.
	
\begin{figure}[t]
	\begin{center}
		\includegraphics[scale=0.30]{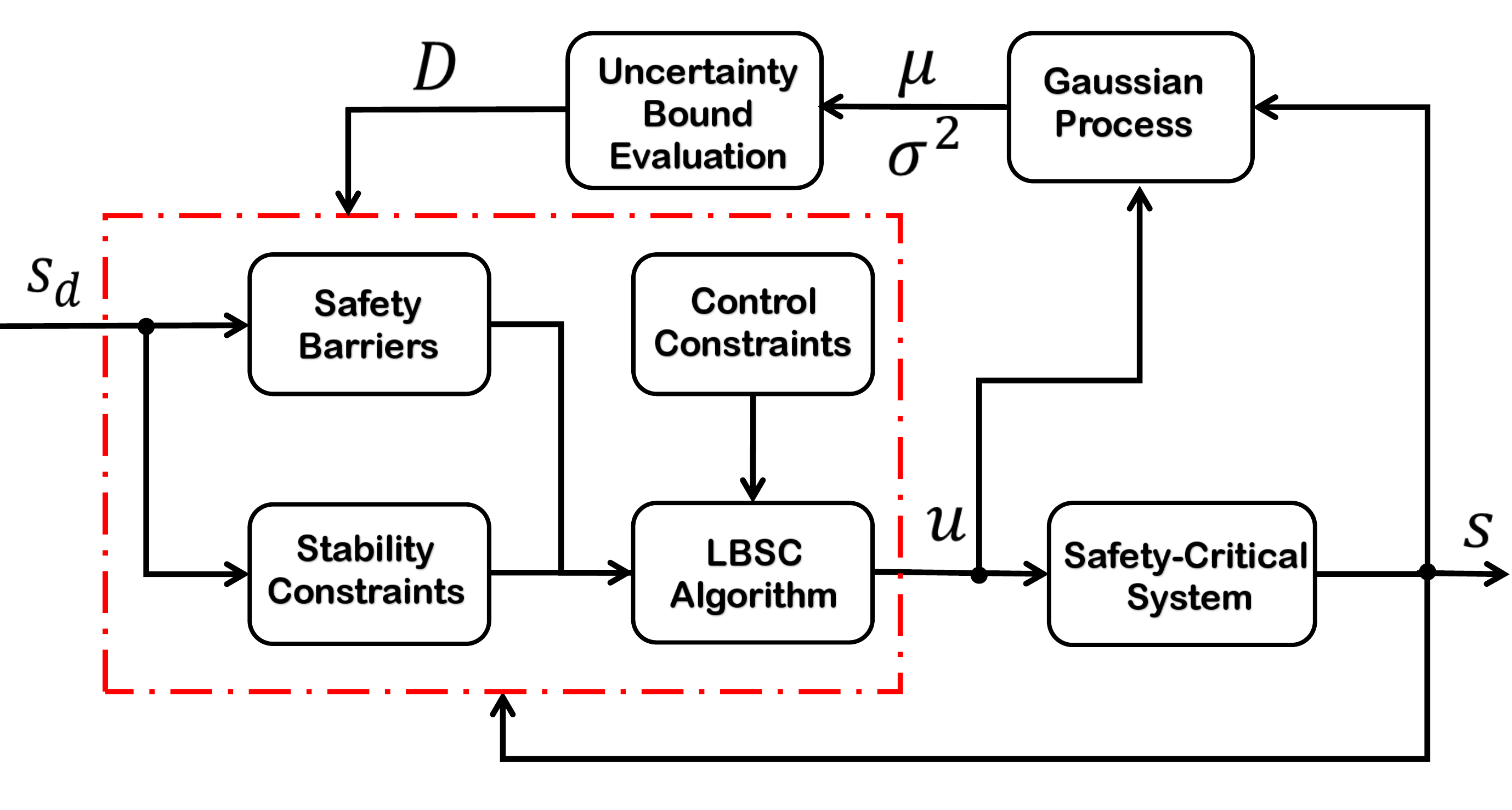}
		\label{diagram}
	\end{center}	\vspace{-6mm}
	\caption{{\bf{Block diagram of our proposed strategy.}}
 GPs are utilized to learn the model errors of the safety-critical systems. A high confidence interval $D$ with respect to the uncertainty bound can be evaluated based on the predicted mean $\mu$ and variance $\sigma^{2}$ of the model errors. Using the high confidence interval $D$, the LBSC is formulated unifying the safety barrier, the stability constraint, and the control constraints. The symbol $s$, $s_{d}$ and $u$  denote the actual system state, the desired system state, and the control input, respectively.}
	\label{fig:illustration}		\vspace{-5mm}
\end{figure}
	
	Considering the limitations of current learning-based MPC approaches and CBF-CLF-QPs methods, learning-based approaches are desirable to be developed to ensure both safety and tracking stability for the nonlinear safety-critical systems with model uncertainties. Particularly, as safety must not be violated, it should mediate the tradeoff between safety and stabilization objective when there exist conflicts between them.

	In this paper, drawing inspiration from the CBF-CLF-QPs framework, we propose a  learning-based algorithm utilizing GPs to model the uncertain dynamical system (inaccurate system parameters, continuous road grade changes, etc) online. The predicted mean and variance of GPs are used to quantify a high confidence interval for uncertainty. With the estimated uncertainty bounds, a safety barrier represented by CBFs is devised to enforce safety, and a stability constraint based on CLFs is formulated to achieve high-performance tracking. Considering input constraints, the learning-based safety-stability-driven control (LBSC) algorithm for safety-critical systems is formulated in a standard QP formulation incorporating the safety barrier and the stability constraint as shown in Fig.~\ref{diagram}.
	
	The $\mathbf{main\ contributions}$ of this paper are presented as follows: 
	\begin{itemize}
		\item We present a novel learning-based algorithm LBSC to achieve safety guarantees and stabilization objectives for the nonlinear safety-critical systems with model uncertainties. The LBSC method is formulated as a constrained QP incorporating safety (represented by CBFs) , stability (represented by CLFs), and control constraints to mediate the tradeoff between safety guarantees and tracking stability.
		
		\item We extend the CBFs and CLFs to the uncertain safety-critical systems, which explicitly uses the estimated uncertainties via GPs to guarantee safety and tracking stability.

		\item We validate the effectiveness of the LBSC algorithm via numerical simulations on the high-speed safety-critical connected cruise control system under model uncertainties.
	\end{itemize}
	
	This paper is organized as follows: 
	Section~\ref{section:preliminaries} presents the preliminaries used in this paper. The problem statement is introduced in Section~\ref{section:problem statement}.
	The proposed LBSC algorithm is presented in Section~\ref{section:method} with necessary derivations. Simulation results are shown in Section~\ref{section:experiment} on a connected cruise control safety-critical system to validate the LBSC approach. Finally, conclusions are drawn in Section~\ref{section:conclusion}.

	\section{Preliminaries}
	\label{section:preliminaries}
	
	In this section, a brief review of Control Barrier Functions (CBFs) and Gaussian process (GPs) are presented.
	
	Consider a nonlinear affine control system
	\vspace{-0.2\baselineskip}
	\begin{equation}
	\dot{x}=f(x)+g(x)u,
	\label{dynamics}	
	\vspace{-0.2\baselineskip}
	\end{equation}
	where $x\in\mathcal{X}\subseteq\mathbb{R}^{n}$, $u\in\mathcal{U}\subseteq\mathbb{R}^{m}$ denote the states and the control input of the system, and the function $f:\mathbb{R}^{n}\rightarrow\mathbb{R}^{n}$ and $g:\mathbb{R}^{n}\rightarrow\mathbb{R}^{n \times m}$ are Lipschitz continuous.

	\subsection{Control Barrier Functions (CBFs)} 
	CBFs have been widely used in control systems to enforce safety constraints~\cite{Ames2014ControlBF}.
	A $\mathit{safety\ set}\ \mathcal{S}$~\cite{Ames2014ControlBF} is considered to motivate a formulation of the CBFs, which is defined by
			\vspace{-0.4\baselineskip}
	\begin{equation}
	\mathcal{S}:=\{x\in\mathbb{R}^{n}|h(x)\geq0\},
	\label{safety set} 
		\vspace{-0.2\baselineskip}
	\end{equation}
    where $h:\mathbb{R}^{n}\rightarrow\mathbb{R}$ is a continuously differentiable function.
	
	\noindent\textbf{Definition 1} (\hspace{-0.05mm}\cite{khalil2002nonlinear})\noindent\textbf{.}
	The set $\mathcal{S}$ is called $\mathit{forward\ invariant}$, if for every $x_{0}\in\mathcal{S}$, $x(t,x_{0})\in\mathcal{S}$ for all $t\in \mathbb{R}_{0}^{+}$.
	
	To ensure forward invariance of the set $\mathcal{S}$, e.g. mobile robots stay in the collision-free safety set at all times, we consider the following definition.
	
	\noindent\textbf{Definition 2} (\hspace{-0.05mm}\cite{Ames2017ControlBF})\noindent\textbf{.}
	For the dynamical system~(\ref{dynamics}), given a set $\mathcal{S}\subset{\mathbb{R}}^{n}$ defined by (\ref{safety set}) for a continuously differentiable function $h:\mathbb{R}^{n}\rightarrow\mathbb{R}$, the function $h$ is called a $\mathit{Zeroing \ Control\  Barrier \  Function}$ $(ZCBF)$ defined on the set $\mathcal{D}$ with $\mathcal{S} \subseteq \mathcal{D} \subset{\mathbb{R}}^{n}$, if there exists an extended class $\mathcal{K}$ function \cite{khalil2002nonlinear} $\alpha$ such that
			\vspace{-0.26\baselineskip}
	\begin{equation}
	\mathop{\sup}_{u \in \mathcal{U}}[L_{f}h(x) + L_{g}h(x)u + \alpha (h(x))]\geq0, \forall x\in\mathcal{D},
			\vspace{-0.25\baselineskip}
	\end{equation}
	where $L$ represents the Lie derivatives. To be more specific:
			\vspace{-0.4\baselineskip}
	\begin{equation}
	L_{f}h(x)=\frac{\partial h(x)}{\partial x}f(x),\ L_{g}h(x)=\frac{\partial h(x)}{\partial x}g(x).
	\label{lie d}
			\vspace{-0.3\baselineskip}
	\end{equation}

	ZCBF is a special control barrier function that comes with asymptotic stability~\cite{Xu2015RobustnessOC}. The existence of a ZCBF implies the asymptotic stability and forward invariance of $\mathcal{S}$ as proved in~\cite{Xu2015RobustnessOC}.

	\subsection{Gaussian Processes}
		\label{section:GP}
	In this paper, we consider an uncertain control affine system with partially uncertain dynamics:\vspace{-1.0mm}
	\vspace{-0.2\baselineskip}
	\begin{equation}
	\dot{x}=f(x)+g(x)u+d(x),
	\label{dynamics_with_d}
	\vspace{-0.2\baselineskip}
	\end{equation}
	where $x\in\mathcal{X}\subseteq\mathbb{R}^{n}$ denotes the system state, $u\in\mathcal{U}\subseteq\mathbb{R}^{m}$ is the control input, functions $f: \mathbb{R}^{n}\rightarrow\mathbb{R}^{n}$ and $g: \mathbb{R}^{n}\rightarrow\mathbb{R}^{n \times m}$ compose a prior model representing our knowledge of the actual system, and model error $d:\mathbb{R}^{n}\rightarrow\mathbb{R}^{n}$ represents uncertain discrepancies between the prior model and the actual system. Similar to~\cite{Wang2017SafeLO}, assume that the functions $f(x)$, $g(x)$ and $d(x)$ in (\ref{dynamics_with_d}) are Lipschitz continuous to generalize the dynamics to the unexplored states.
	
	A GP is an efficiently nonparametric regression method to estimate complex functions and their uncertain distribution~\cite{Rasmussen2005GaussianPF}.
	GPs can be used to learn the model error $d(x)$ using the collected data from the system during operation. To make the problem tractable, similar to Assumption 1 in~\cite{Berkenkamp2016SafeLO}, the following regularity assumption is considered. 

	\noindent\textbf{Assumption}\noindent\textbf{.}
	The unknown model error $d(x)$ has a bounded norm in the associated Reproducing Kernel Hilbert Space (RKHS)~\cite{scholkopf2002learning}, corresponding to a differentiable kernel $k$. 
	\label{regularity_assumption}
	
	This assumption can be interpreted as a requirement on the smoothness of the model error $d(x)$ (e.g. inaccurate system parameters, continuous road grade changes for a car). Also, its boundedness implies that $d(x)$ is regular with respect to the kernel~\cite{srinivas2012information}.
	In this study, GPs~\cite{Rasmussen2005GaussianPF} are used to learn the model uncertainty $d(x)$ in (\ref{dynamics_with_d}) with the regularity assumption.
	
	 Given $n$ observations $\mathbf{D}_{n}:=\{x_{i},\hat{d}(x_{i})\}^{n}_{i=1}$, the mean and variance of $d(x_{*})$ at the query state $x_{*}$ can be given by
	\vspace{-1mm}
	\begin{equation}
	\mu(x_{*})=\mathbf{k}_{n}^{T}(\mathbf{K}+\sigma^{2}\mathbf{I})^{-1}\hat{\mathbf{d }}_{n},
	\label{mean}
		\vspace{-1mm}
	\end{equation}
	\begin{equation}
	\sigma^{2}(x_{*})=k(x_{*},x_{*})-\mathbf{k}_{n}^{T}(\mathbf{K}+\sigma^{2}\mathbf{I})^{-1}\mathbf{k}_{n},
	\label{var}
	\vspace{-1.0mm}
	\end{equation}
	respectively, where $\hat{\mathbf{d}}_{n}=[\hat{d}(x_{1}),\hat{d}(x_{2}),...,\hat{d}(x_{n})]$ is the observed vector subject to a zero mean Gaussian noise $\omega \sim\mathcal{N}(0,\sigma^{2})$. $\mathbf{K}\in\mathbb{R}^{n\times n}$ is the covariance matrix with entries, where $[\mathbf{K}]_{(i,j)}=k(x_{i},x_{j})$, $i,j \in \left\{ 1, ..., n \right\}$, and $k(x_{i},x_{j})$ is the kernel function. $\mathbf{k}_{n}=[k(x_{1},x_{*}),k(x_{2},x_{*}),...,k(x_{n},x_{*})]$, and $\mathbf{I}\in\mathbb{R}^{n\times n}$ is the identity matrix.
	
	With the system model error $d(x)$ learned by the GPs, a high probability confidence interval $\mathcal{D}(x)$ on the uncertain dynamics $d(x)$ can be obtained to enforce its uncertainty bound by designing the constant $c_{\delta}$~\cite{Berkenkamp2016SafeLO}. 
		\vspace{-0.3\baselineskip}
	\begin{equation}
	\mathcal{D}(x)=\{d\ |\ \mu(x)-c_{\delta}\sigma(x) \leq d \leq \mu(x)+c_{\delta}\sigma(x)\}.
	\label{high confidence interval}
		\vspace{-0.3\baselineskip}
	\end{equation}
	where $c_{\delta}$ is a design parameter to get $(1 - \delta)$ confidence, $\delta \in (0, 1)$.
	
	For instance, $95.5\%$ and $99.7\%$ confidence of the uncertainty bound can be achieved at $c_{\delta}=2$ and $c_{\delta}=3$, respectively.	
	\vspace{-0.3\baselineskip}
	\section{Problem Statement}	
	\vspace{-0.3\baselineskip}
	\label{section:problem statement}
	Consider the uncertain nonlinear control affine system (\ref{dynamics_with_d}) with a given initial state. We aim to design a learning-based controller to track the desired trajectory. The controller must satisfy both the initial state constraints and control input constraints. Meanwhile, it should rapidly drive the system (\ref{dynamics_with_d})
	to the target state with safety guarantees under model uncertainties. Specifically, the following desired objectives need to be satisfied:   
	
	\begin{itemize}
		\item \textbf{Safety}: The control scheme should guarantee the safety of the uncertain safety-critical systems under model uncertainties with a probability that can be made arbitrarily high through the controller design.
		\item \textbf{Tracking stability}: For feasible desired trajectories, the tracking errors can rapidly converge to a sufficiently small neighborhood of the desired state with a probability that can be made arbitrarily high through the controller design.
		\item \textbf{Adaptability}: The controller should enable the closed-loop control system to continuously adapt to online changes of system parameters as well as continuous environment disturbances.
		\item \textbf{The tradeoff between safety and stability constraints:} The controller can mediate the tradeoff between safety guarantees and high-performance tracking with respect to tracking accuracy and convergence rate enforced by stability constraints.
	\end{itemize}\vspace{-1.5mm}

	\section{Learning-based Safety-Stability-Driven Control (LBSC) Approach}
	\label{section:method}
	In this section, inspired by the CBF-CLF-QPs~\cite{Ames2019ControlBF}, an LBSC controller is proposed based on GPs under the Lipschitz continuous assumption for the system~(\ref{dynamics_with_d}) and the \textbf{Assumption} in \ref{section:GP}. The performance of the LBSC algorithm is analyzed with respect to four control objectives addressed in~\ref{section:problem statement}. Moreover, the safety guarantees and tracking stability of the LBSC algorithm will be formulated mathematically in detail.	\vspace{-1mm}
	\subsection{Safety Barrier}
	\label{rsafetyc}
	Based on the ZCBF defined in \noindent\textbf{Definition 2}, a safety barrier for safety-critical systems can be constructed. Concretely, we aim to design a safety barrier for the uncertain system to keep the state $x$ in the forward invariant safety set. 
	
    As the model error $d(x)$ is unknown in prior, GPs are employed to estimate the model uncertainties in terms of the predicted mean $\mu(x)$ and variance $\sigma(x)$ of the model error $d(x)$ through (\ref{mean})and~(\ref{var}). 
	
	Using the definition of the ZCBF and the high confidence interval $D(x)$ defined by (\ref{high confidence interval}) with respect to model uncertainties, for all $x\in\mathcal{S}$, the following safe control space $K_{rzbf}$ is formulated for the uncertain dynamical system (\ref{dynamics_with_d}).
	\vspace{-0.25\baselineskip}
	\begin{equation}
	K_{rzbf}(x) = \{u\in\mathcal{U}| 
	\mathop{\inf}_{d \in \mathcal{D}(x)}[\dot{h}(x) +\alpha (h(x))]\geq 0\}.
	\label{k_rzbf}
	\vspace{-0.5\baselineskip}
	\end{equation}
	where $h(x)$ is a ZCBF, $\dot{h}(x) = \frac{\partial h(x)}{\partial x}\dot{x}=L_{f}h(x) + L_{g}h(x)u + L_{d}h(x)$, and $L_{d}h(x)$ denotes the Lie derivatives of $h$ with respect to the model error $d(x)$. 
	
	\noindent\textbf{Lemma 1}\noindent\textbf{.}
	Given a set $\mathcal{S}\subset \mathbb{R}^{n}$ defined by (\ref{safety set}) with an associated ZCBF $h(x)$, the control input $u \in K_{rzbf}$ has a probability of at least $(1-\delta)$, $\delta \in (0,1)$, to guarantee the forward invariance of the set $\mathcal{S}$ for the uncertain dynamical system (\ref{dynamics_with_d})
	
	\begin{proof}
		From (\ref{high confidence interval}), there is a probability of at least $(1-\delta)$ such that the bounded model uncertainty $d(x) \in \mathcal{D}(x)$ for all $x\in\mathcal{X}$. Since the control input $u\in\mathcal{U}$ of the safe control space $K_{rzbf}$ satisfies the constraint in (\ref{k_rzbf}), the following result holds with a probability of at least $(1-\delta)$:	\vspace{-0.5\baselineskip}
		\begin{equation}
		\dot h(x) + \alpha(h(x)) \geq 0, \forall x\in\mathcal{S}.
		\vspace{-0.5\baselineskip}
		\end{equation}
		
		As a result, the control input $u\in\mathcal{U}$ of the safe control space $K_{rzbf}$ has a probability of at least $(1-\delta)$ 
		to guarantee the forward invariance of the set $\mathcal{S}$ for the uncertain dynamical system~(\ref{dynamics_with_d}) as proven in~\cite{Ames2017ControlBF}.
	\end{proof}
	
	For convenience, the constraint in (\ref{k_rzbf}) can be equivalently expressed as
						\vspace{-2.5mm}
	\begin{equation}
	L_{f}h(x) + L_{g}h(x)u + L_{\mu}h(x) -c_{\delta}|L_{\sigma}h(x)|  \geq  -\alpha(h(x)),
	\label{rcbf constraints}
						\vspace{-1.0mm}
	\end{equation}
	where $L_{\mu}h(x)$ and $L_{\sigma}h(x)$ denote the Lie derivatives of $h(x)$ with respect to $\mu$ and $\sigma$, respectively.
	
	\noindent\textbf{Remark 1}\noindent\textbf{.}
	Using more informative data collected for the system dynamics, the bounded uncertainty $\sigma$ will gradually decrease. Thus, the probability rendering $\mathcal{S}$ forward invariant is much higher than $(1-\delta)$ in most cases.
	
	\subsection{Stability Constraint}
	In this study, a stability constraint is developed to enforce rapidly tracking stability with respect to exponential stability~\cite{khalil2002nonlinear} in the presence of model uncertainties.
	
	The desired stable high-performance tracking with respect to convergence rate and tracking accuracy can be captured by an $exponentially$ $stabilizing$ $control$ $Lyapunov$ $function$ $(ES-CLF)$~\cite{AmesRapidly}.
	Based on an ES-CLF and the high confidence interval $D(x)$ defined by (\ref{high confidence interval}) with respect to model uncertainties, for all $x\in\mathcal{S}$, we consider the following admissible stabilizing control space to ensure stable high tracking performance for the uncertain dynamical system (\ref{dynamics_with_d}) 
	\vspace{-0.5\baselineskip}
	\begin{equation}
	K_{rclf}(x) = \{u\in\mathcal{U}| 
	\mathop{\sup}_{d \in \mathcal{D}(x)}[\dot{V}(x) + c V(x)]\leq 0\},
	\label{prclf}
	\vspace{-0.5\baselineskip}
	\end{equation}
	where $V(x)$ is an ES-CLF, $\dot{V}(x) = \frac{\partial V(x)}{\partial x}\dot{x}=L_{f}V(x) + L_{g}V(x)u + L_{d}V(x)$, and c is a positive constant.
	
	Similar to part~\ref{rsafetyc}, the uncertainty bound of $d(x)$ in (\ref{dynamics_with_d}) can be estimated via GPs.
	
	\noindent\textbf{Lemma 2}\noindent\textbf{.}
	Given a set $\mathcal{S}\subset \mathbb{R}^{n}$ defined by (\ref{safety set}) with an associated ES-CLF, the control input $u\in K_{rclf}$ has a probability of at least $(1-\delta)$, $\delta \in (0,1)$, to exponentially stabilize the uncertain dynamical system (\ref{dynamics_with_d}).
	
	\begin{proof}
		From (\ref{high confidence interval}), there is a probability of at least $(1-\delta)$ such that the bounded model uncertainty $d(x) \in \mathcal{D}(x)$ for all $x\in\mathcal{X}$. 
		Since the control input $u\in\mathcal{U}$ of the stabilizing control space $K_{rclf}$ satisfies the constraint in (\ref{prclf}), the following result holds with a probability of at least $(1-\delta)$:
		\vspace{-0.4\baselineskip}
		\begin{equation}
		\dot V(x) + cV(x) \leq 0, \forall x\in\mathcal{S}.
		\vspace{-0.4\baselineskip}
		\end{equation}
		
		Hence, the control input $u\in\mathcal{U}$ of the control space $K_{rclf}$ has a probability of at least $(1-\delta)$ to exponentially stabilize the uncertain dynamical system~(\ref{dynamics_with_d}) as shown in~\cite{AmesRapidly}.
	\end{proof}
	
	For convenience, the constraints in~(\ref{prclf}) can be simplified as follows, 
		\vspace{-0.3\baselineskip}
	\begin{equation}
	L_{f}V(x) + L_{g}V(x)u + L_{\mu}V(x)+c_{\delta}|L_{\sigma}V(x)| \leq -c V(x).
	\label{rclf constraints}
		\vspace{-0.3\baselineskip}
	\end{equation}
	
	With an associated ES-CLF, we can enforce the state of the uncertain dynamical system (\ref{dynamics_with_d}) to rapidly converge to the desired state with a probability larger than $(1-\delta)$ based on~(\ref{rclf constraints}).
	
	\noindent\textbf{Remark 2}\noindent\textbf{.}
	With more informative data collected for the system dynamics, the bounded uncertainty $\sigma$ will gradually decrease. Thus, the probability of the control input 
	$u\in\mathcal{U} \in K_{rclf}$ to enforce
	tracking stability with respect to exponential stability for the uncertain dynamical system (\ref{dynamics_with_d}) is much higher than $(1-\delta)$ in most cases.\vspace{-2mm}
	\subsection{Learning-based Safety-Stability-Driven Controller (LBSC)}
	In practical applications, the safety-critical systems are usually subject to control input constraints, such as torque saturation constraints. In this study, using the estimated uncertainty bound estimated by GPs, the LBSC approach is formulated as a quadratic program (QP) controller incorporating the safety barrier (represented by CBF) in (\ref{rcbf constraints}), stabilization objective (represented by ES-CLF) in (\ref{rclf constraints}) and control constraints. 
	
 Furthermore, the tradeoff of safety and tracking stability is mediated by setting relaxation variables for safety and stability constraints. The proposed LBSC controller is then formulated as follows:\vspace{-1mm}
	\begin{alignat}{2}
	\label{LBSC}
	u^{*}(x)=\mathop{\arg\min}_{u \in {\mathbb{R}^{m}},(\varepsilon,\eta) \in {\mathbb{R}}} &\frac{1}{2}u^{T}H(x)u +K_{\varepsilon}\varepsilon^{2}+K_{\eta}\eta^{2}, \\
	\mbox{s.t.} \quad
	A_{rzbf}&u+b_{rzcbf} \leq \varepsilon,\tag{Safety} \\
	A_{rclf}&u+b_{rclf} \leq \eta, \tag{Stability} \\
	u_{min}&\leq u \leq u_{max},\tag{Control Constraints}
	\end{alignat}\vspace{-2mm}	where 
	\begin{alignat}{2}
	&A_{rzcbf} =-L_{g}h(x), \notag \\
	&b_{rzcbf}= -L_{f}h(x) - L_{\mu}h(x)+c_{\delta}|L_{\sigma}h(x)| - \alpha(h(x)), \notag \\
	&A_{rclf} =L_{g}V(x), \notag \\
	&b_{rclf}= L_{f}V(x) + L_{\mu}V(x) +c_{\delta}|L_{\sigma}V(x)|+ c V(x),\notag  \vspace{-4mm}
	\end{alignat}
	$u_{min}\in\mathcal{U}$ and $u_{max}\in\mathcal{U}$ are the lower bound and the upper bound of the control inputs, respectively; $H(x)\in\mathbb{R}^{m\times m}$ is positive definite; c is a positive constant; $\varepsilon$ and $\eta$ are relaxation variables for safety and stability constraints, respectively; $K_{\epsilon}$ and $K_{\eta}$ are two positive constants to penalize safety and tracking stability violation, respectively.
	
	The solution $u^{*}(x)$ to the QP problem in (\ref{LBSC}) is always feasible because the relaxation variables $\varepsilon$ and $\eta$ can ensure no conflict among the safety, stability, and control input constraints. Similar to Theorem 3 in~\cite{Xu2015RobustnessOC}, it can be proved that $u^{*}(x)$ is Lipschitz continuous. Moreover, the optimization is not sensitive to the parameters $K_{\varepsilon}$ and $K_{\eta}$ as long as they are large enough (e.g. $K_{\varepsilon}=10^{30}$, $K_{\eta}=10^{20}$). In this way, the safety and stability constraints violations will be penalized heavily when the parameters are set large values. 
	
	\noindent\textbf{Remark 3}\noindent\textbf{.}
	Note that $K_{\varepsilon}$ (related to safety) is set extremely larger than $K_{\eta}$ (related to stabilization) to make the safety guarantees much stricter than the stabilization objective of tracking. As a result, the QP formation in (\ref{LBSC}) can handle the tradeoff between safety and tracking stability. If no control inputs are satisfying both safety and tracking stability constraints, it would sacrifice tracking performance to guarantee safety constraints.
	\section{Experiments}
	\label{section:experiment}
In this section, the connected cruise control (CCC) system, a typical nonlinear safety-critical system, is examined to illustrate the performance of the proposed LBSC approach. For a CCC system consisting of a team of cars (one controlled car among several human-driven cars), urgent deceleration of the lead car in unexpected situations may cause the car behind to collide with the front car. Moreover, the self-driving car in the CCC system may drive in unsafe states due to model uncertainties. It poses a critical challenge to achieve accurate tracking and safety guarantees for the CCC system in these situations. The LBSC approach is investigated by comparing it against the other three baselines in simulations for the CCC system.
	
	\begin{itemize}
		\item \textbf{GP-based Adaptive sampling (GPAS) Method}~\cite{Teck2018GaussianPA}:
		An adaptive sampling based MPC strategy through GPs and cross-entropy. The GPAS is performed for the CCC system by elaborating a cost function measuring the tradeoff between safety and tracking performance.
		\item \textbf{Control barrier function and control Lyapunov function based quadratic programs (CBF-CLF-QPs)}~\cite{Ames2017ControlBF}: The ZCBF is utilized for the CBF-CLF-QPs~\cite{Ames2017ControlBF} as it provides robustness property under model perturbations as investigated in~\cite{Xu2015RobustnessOC}.
		\item \textbf{LBSC-N}: An ablation version of the proposed LBSC algorithm in (\ref{LBSC}). In the LBSC-N the value of the weight $K_{\epsilon}$ for safety constraints is equal to the weight $K_{\eta}$ for stability constraints, i.e., $K_{\eta}=K_{\varepsilon}=10^{30}$.
	\end{itemize}

	\subsection{Experimental Setup}
	For the CCC system, we consider a chain of five cars following in order on a straight road with uncertain slope or grade, including autonomous car 4 and four human-driven cars including cars 1, 2, 3, and 5. Car 1 leads the chain of five cars with a series of aggressive accelerations and urgent braking events. To form the fleet of five cars, the space headway between a car and its front car is required to range from the minimum safe distance to the maximum tracking distance. With this constraint of the space headway, a car can guarantee safety while rapidly tracking the car in front of it in the presence of urgent deceleration and acceleration.
	In the CCC system, assume that each car can access its front cars’ positions, velocities, and accelerations via vehicle-to-vehicle
	(V2V) communication~\cite{He2018DatabasedFO}.

	For the CCC system, its states are denoted as $q = [p_{1}$, $v_{1} $, $a_{1}$, $p_{2}$, $ v_{2}$, $a_{2}$, $p_{3}$, $ v_{3}$, $a_{3}$, $p_{4}$, $v_{4}$, $ a_{4}$, $p_{5}$, $v_{5}$, $a_{5}]$, where $p_{i}$, $v_{i}$, and $a_{i}$ are the position(in $m$), velocity (in $m/s$), acceleration (in $m/s^{2}$) of the $i$th car, respectively; and the control inputs are represented as $u =[u_{1}$, $ u_{2}$, $ u_{3}$, $ u_{4}$, $u_{5}]$, where the control input $u_{i}$ (in Newtons) of the $i$th car is the total wheel force.

	The car dynamics are inspired by work~\cite{Xu2015RobustnessOC}:
	\begin{equation}
	\left[
	\begin{array}{cccc}
	\dot{p}_{i}\\ 
	\dot{v}_{i}
	\end{array}
	\right]
	=
	\left[
	\begin{array}{cccc}
	{v}_{i}\\ 
	{a}_{i}
	\end{array}
	\right]
	=
	\left[
	\begin{array}{cccc}
	v_{i}\\
	\frac{-F _r -F _f}{M}
	\end{array}
	\right ]
	+\left[
	\begin{array}{cccc}
	0\\ 
	g\Delta\theta
	\end{array}
	\right]
	+\left[
	\begin{array}{cccc}
	0\\ 
	\frac{1}{M}
	\end{array}
	\right]
	u_{i},
	\label{car dynamics}
	\end{equation}
	where $F_{r} = f_{0} + f_{1}v_{f}+ f_{2}v_{f}^{2}$ is the aerodynamic drag (in Newtons) with constants $f_{0},\ f_{1}$, and $f_{2}$; $F_{f} = f_{f}Mg$ is the rolling resistance (in Newtons), and $f_{f}$ is the rolling resistance coefficient determined empirically; $M$ is the mass of the car, and $g$ is the gravitational acceleration; $\Delta\theta$ is a perturbation to $v_{i}$ (reflecting unmodeled road grade).
	
	\begin{table}[htbp]
		\caption{Accurate System Parameter Values in Simulations}
		\vspace{-2.5mm}
		\label{parameter}
		\begin{tabular}{cccc}
			\midrule  
			\midrule 
			$M$&1650 $kg$ &	$g$&9.81 $m/s^{2}$\\
			$k_{b}$& 30 & $k_{p}$& 2000\\
			$f_{f}$& 0.015 & $f_{0}$&0.1 $N$\\
			$f_{1}$& 5 $N \cdot s/m$ & $f_{2}$&0.25 $N\cdot s^{2}/m^{2}$\\
			$v_{max}$&40 $m/s$ &$a_{max}$& 0.3 $\times$ 9.81 $m/s^{2}$\\
			$c_{a}$& 0.3 &$c_{d}$& 0.3\\
			\midrule  
			\midrule  
		\end{tabular}
		\vspace{-3mm}
		\centering
	\end{table}
	
	On the other hand, the control input $u_{i}$ considering in (\ref{car dynamics}) for $ith$ human-driven cars are described as follows~\cite{He2018DatabasedFO}: \vspace{-2mm}
	\begin{equation}
	\begin{aligned} 
	\label{ccc systems}
	u_{i}(t) =&k_{b}(V_{i}(p_{i+1}(t)- p_{i}(t)-l_{i})-v_{i}(t))\\&+ k_{p}(v_{i+1}(t) - v_{i}), \\
	\end{aligned}  \vspace{-1.5mm}
	\end{equation}
	\begin{equation}
	\label{control law}
		\vspace{-0.15\baselineskip}
	V_{i}(B_{i})=\begin{cases}
	0&\text{if $B_{i}\leq B_{st,i}$}\\
	k_i(B_{i}- B_{st,i})&\text{if $B_{st,i} < B_{i} < B_{go,i}$}\\
	v_{max}&\ B_{i}\geq B_{go,i}
	\end{cases},
	\vspace{-0.05\baselineskip}
	\end{equation}
	where $k_{b}$ and $k_{p}$ denote two control gains; $B_{i}=p_{i+1}-p_{i}-l_{i}$, and $l_{i}$ denotes the length of $ith$ car; $B_{st,i}$ and $B_{go,i}$ denote coefficients of the control law used by the $ith$ human-driven car, $i=2,3,5$. For a small following distance ($B_{i} < B_{go,i}$), the car intends to stop, while for a large following distance ($B_{i}\geq B_{go,i}$), it intends to travel with the speed limit $v_{max}$.
	In our experiments, $B_{st,i} = 25m$ and $B_{go,i} = 100m$ are determined empirically.

	Other parameter values of the CCC systems are set referred to~\cite{Ames2014ControlBF} and~\cite{He2018DatabasedFO}.
	
	In the simulations, the accurate system parameters are given in Table~\ref{parameter}. The initial states of the CCC system are set as $q = [\ 0 \ m$, $18$ $m/s$ , $0 m/s^{2}$, $60 \  m$, $18 $ $m/s$, $0 m/s^{2}$, $120 \ m$, $18$ $m/s$, $0\ m/s^{2}$, $180\ m$, $18m/s$,  $0 m/s^{2}$, $240 \ m$, $18$ $m/s$, $0\ m/s^{2}$$\ ]$.

In the presence of model uncertainties (i.e., uncertain rolling resistance, aerodynamic drag, and uncertain slope or grade), we aim to control the input of the autonomous car 4 between the human-driven car 3 and car 5 to enable the autonomous car 4 to rapidly converge to a fixed cruising velocity, maintain it and meanwhile keep a safe space headway (between 25 $m$ and 100 $m$) with its front human-driven car 3 in the chain of five cars.

\begin{figure}	
		\centering
		\hspace{-7mm}
		\subfigure[]{
			\label{fig:car1 case1}
			\includegraphics[scale=0.265]{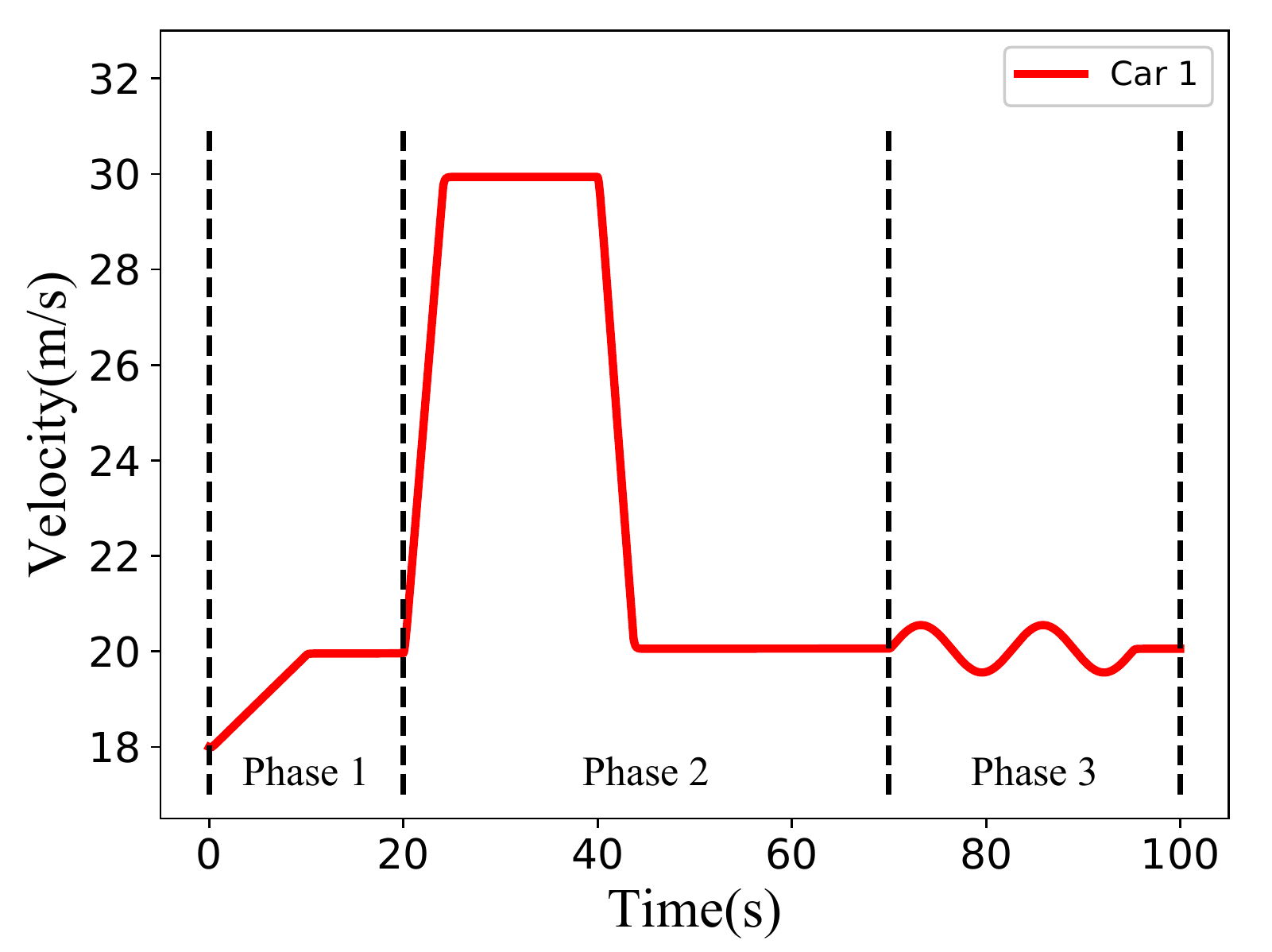}
		}\hspace{-3mm}
		\subfigure[]{
			\label{fig:car1 case2}
			\includegraphics[scale=0.265]{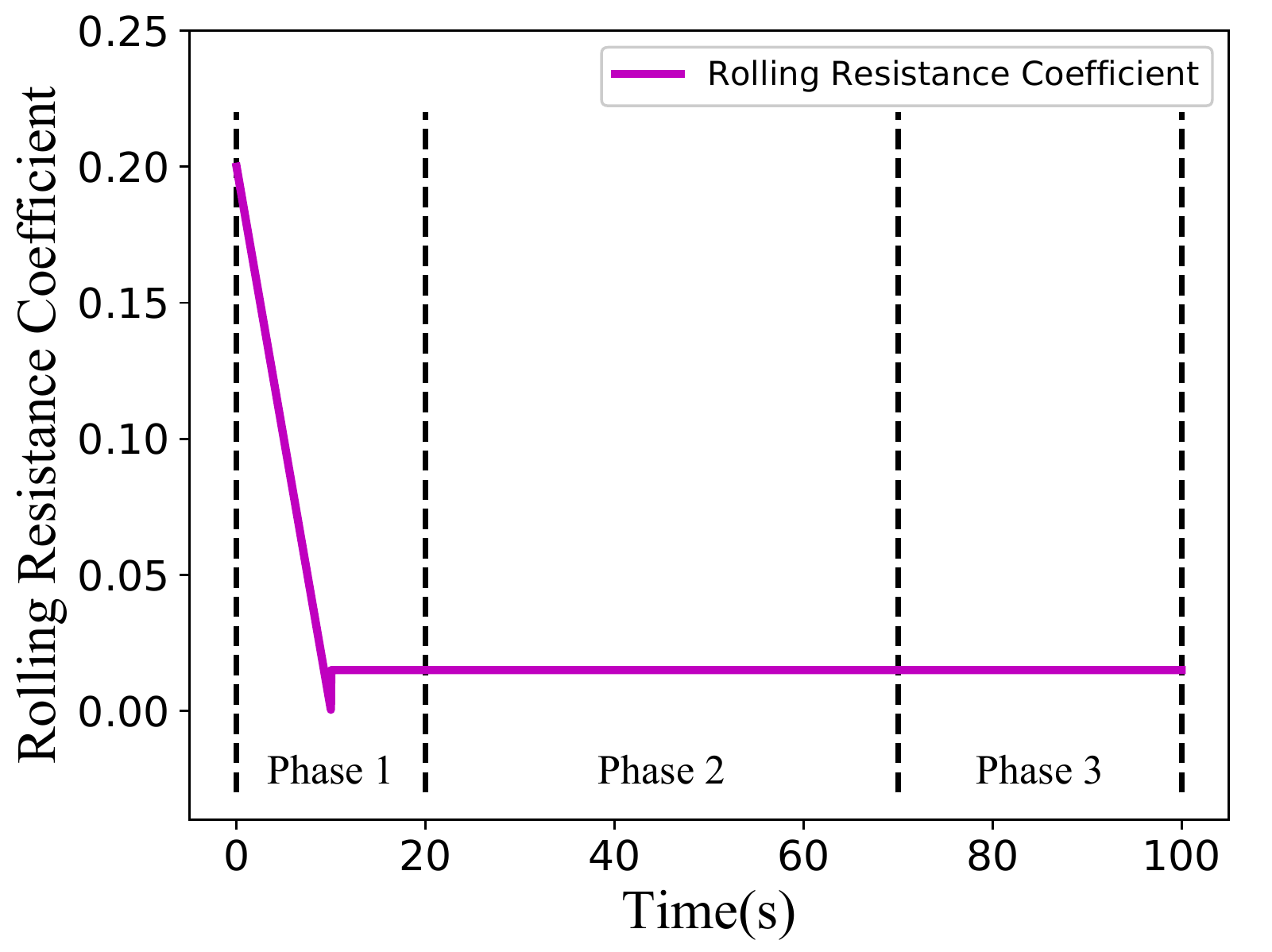}
		}\hspace{-7mm}
		\vspace{-2mm}
		\caption{(a) The velocity of the lead car in three phases. (b) The rolling resistance coefficient of the road.}
		\vspace{-4mm}
		\label{fig:ccc_vel}	\vspace{-2mm}
	\end{figure}

To study the tracking performance of the LBSC controller in various situations, the velocity of the lead car varies according to the three phases as shown in Fig.~\ref{fig:car1 case1}. 

\noindent\textbf{Phase 1 (0$s\sim$20$s$)}\noindent\textbf{.} 
The lead car 1 accelerates from $18m/s$ to $20m/s$ with a fixed acceleration and then drives smoothly at a constant speed of $v_{d}$.

\noindent\textbf{Phase 2 (20$s\sim$70$s$)}\noindent\textbf{.} 
The lead car 1 rapidly accelerates to reach the velocity of $30m/s$ and then maintains a fixed speed until the $40s$. After that, it urgently decelerates back to the velocity of $20m/s$.

\noindent\textbf{Phase 3 (70$s\sim$100$s$)}\noindent\textbf{.} 
The velocity of the lead car 1 is set to slightly and continuously varying as a $sin$ function.

\noindent\textbf{Road Disturbances}\noindent\textbf{.}
In each phase, the rolling resistance coefficient $f_{f}$ of the road is shown in Fig.~\ref{fig:car1 case2}. In phase 3, A grade perturbation is given by $g \Delta \theta = 2.5sin(0.5t)$ when the simulation time $t \geq 70s$. 

	\noindent\textbf{Model Uncertainties}\noindent\textbf{.} Assume that the autonomous car 4 has prior knowledge of nominal models of its dynamics and the other human-driven cars. To illustrate the discrepancies between the nominal and the accurate models, the parameter values of the nominal models known by car 4 in prior are set differently from the accurate values. 
	\begin{itemize}
		\item The parameters of the crude model of car 4 in (\ref{car dynamics}) are set as $f_{f} = 0.2$, $F_{r} = 0$, $\Delta \theta = 0$.
		\item The parameters of the other human-driven cars in (\ref{control law}) and (\ref{car dynamics}) are set as $k_{b} = 20$, $k_{p} = 1000$, $f_{f} = 0.2$, $F_{r} = 0$, $\Delta \theta = 0$.
	\end{itemize}
	\noindent\textbf{Safety Constraints}\noindent\textbf{.}
	The position of the autonomous car 4 is constrained within an interval between the $B_{st,4}$ and $B_{go,4}$,  
		\vspace{-4mm}
	\begin{equation}
	B_{st,4}  \leq p_{3} - p_{4} \leq B_{go,4}. 
		\vspace{-2mm}
	\end{equation}

	This safety range aims to maintain a safe following distance as specified by a space headway. In the experiments, we set $B_{st,4}=25m$, $B_{go,4}=100m$. The space headway hence ranges from 25$m$ to 100$m$. The barrier function $h_{1}(x)$ and $h_{2}(x)$ can be parameterized as
		\vspace{-2mm}
	\begin{equation}
	\begin{split}	
	h_{1}(x) = p_{4}-p_{3}-B_{st,4},\\
	h_{2}(x) = -p_{4}+p_{3}+B_{go,4},
	\end{split}
	\vspace{-2mm}
	\end{equation}
	respectively. In the following experiments, we set $B_{st,4}=25m$, $B_{go,4}=100m$.
	
	\noindent\textbf{Tracking Performance}\noindent\textbf{.} The LBSC controller aims to rapidly drive the autonomous car at the desired cruising velocity with high tracking accuracy. This stabilization objective is encoded as an ES-CLF in a standard quadratic form as:
			\vspace{-0.4\baselineskip}
	\begin{equation}
	V(x) = \frac{1}{2}||v_{4}-v_{des4}|| ^2.
		\vspace{-0.1\baselineskip}
	\end{equation}
	
	In each phase, the tracking errors are evaluated via Mean Absolute Error (MAE) between the desired and tracked velocity of the autonomous car 4:
	\vspace{-2mm}
	\begin{equation}
				\vspace{-0.1\baselineskip}
	MAE=\frac{1}{T}\sum_{t=1}^{T}\parallel v_{4}(t)-v_{des}(t)\parallel 
	\label{MAE}
		\vspace{-0.1\baselineskip}
	\end{equation}
	where $T$ is the number of samples in each phase.
	
	\noindent\textbf{Control Constraints}\noindent\textbf{.} 
	Similarly to~\cite{Ames2014ControlBF}, the control input set considered in the CCC system is defined by:
			\vspace{-0.15\baselineskip}
	\begin{equation}
	U_{CCC}= [u_{min},\ u_{max}] = [-c_{d}Mg,\ c_{a}Mg], 
			\vspace{-0.15\baselineskip}
	\end{equation}
	where $u_{min}$ and $u_{max}$ are the maximum control value for deceleration and acceleration, respectively. $c_{d}$ and $c_{a}$ denote deceleration and acceleration coefficients, respectively.
	
	In the following experiments, each GPs model uses the past $T_{s}=30$ observations collected at 50 $Hz$ to predict model uncertainties and road perturbations online. To generate high confidence intervals in~(\ref{high confidence interval}), we use $\mathcal{D}(x)=\{d\ |\ \mu(x)-c_{\delta}\sigma(x) \leq d \leq \mu(x)+c_{\delta}\sigma(x)\}$, where $c_{\delta} = 3$.
	The GPs are implemented and tuned using the Python library scikit-learn~\cite{Pedregosa2011ScikitlearnML}.
	
	The feedback controller $u(x)$ for the autonomous car 4 can then be obtained by the following QP problem: \vspace{-1mm}
	\begin{alignat}{2}
	\label{LBSC_CCC}
	u^{*}(x)=\mathop{\arg\min}_{u \in {\mathbb{R}^{1}},(\varepsilon,\eta) \in {\mathbb{R}}} \frac{1}{2}u^{T}H&(x)u +K_{\varepsilon}\varepsilon^{2}+K_{\eta}\eta^{2}, \\
	\mbox{s.t.} \quad
	A_{1}u+b_{1} \leq& \varepsilon,\ A_{2}u+b_{2} \leq \varepsilon, \tag{Safety} \\
	A_{3}u+b_{3} \leq& \eta, \tag{Stability} \\
	u_{min} \leq u \leq& u_{max},\tag{Control Constraints} 
	\vspace{-1mm}
	\end{alignat}
	where $u$ denotes the control inputs. $H(x)=M^{-2}$, $K_{\epsilon}=10^{30}$, $K_{\eta}=10^{20}$, $c_{\delta}=3$. \vspace{-3mm}
	\begin{alignat}{2}
	&A_{1} =-L_{g}h_ 1(x), A_{2} =-L_{g}h_ 2(x), \notag \\
	&b_{1}= -L_{f}h_ 1(x) - L_{\mu}h_ 1(x)+c_{\delta}|L_{\sigma}h_ 1(x)| - \alpha(h_ 1(x)), \notag\\
	&b_{2}= -L_{f}h_ 2(x) - L_{\mu}h_ 2(x)+c_{\delta}|L_{\sigma}h_ 2(x)| - \alpha(h_ 2(x)), \notag   \\
	&A_{3} =L_{g}V(x), \notag \\
	&b_{3}= L_{f}V(x) + L_{\mu}V(x) +c_{\delta}|L_{\sigma}V(x)|+ cV(x),\notag
	\end{alignat}
	where the corresponding extended class $\mathcal {K}$ function $\alpha$ is simply chosen as $\alpha(h_{1}) = 5h_{1}$ and $\alpha(h_{2}) = 5h_{2}$, the positive constant is set as $c = 0.6$. 

	\begin{figure}[t]		
		\centering
		\hspace{-7mm}
		\subfigure[]{
			\label{fig:err1}
			\includegraphics[scale=0.265]{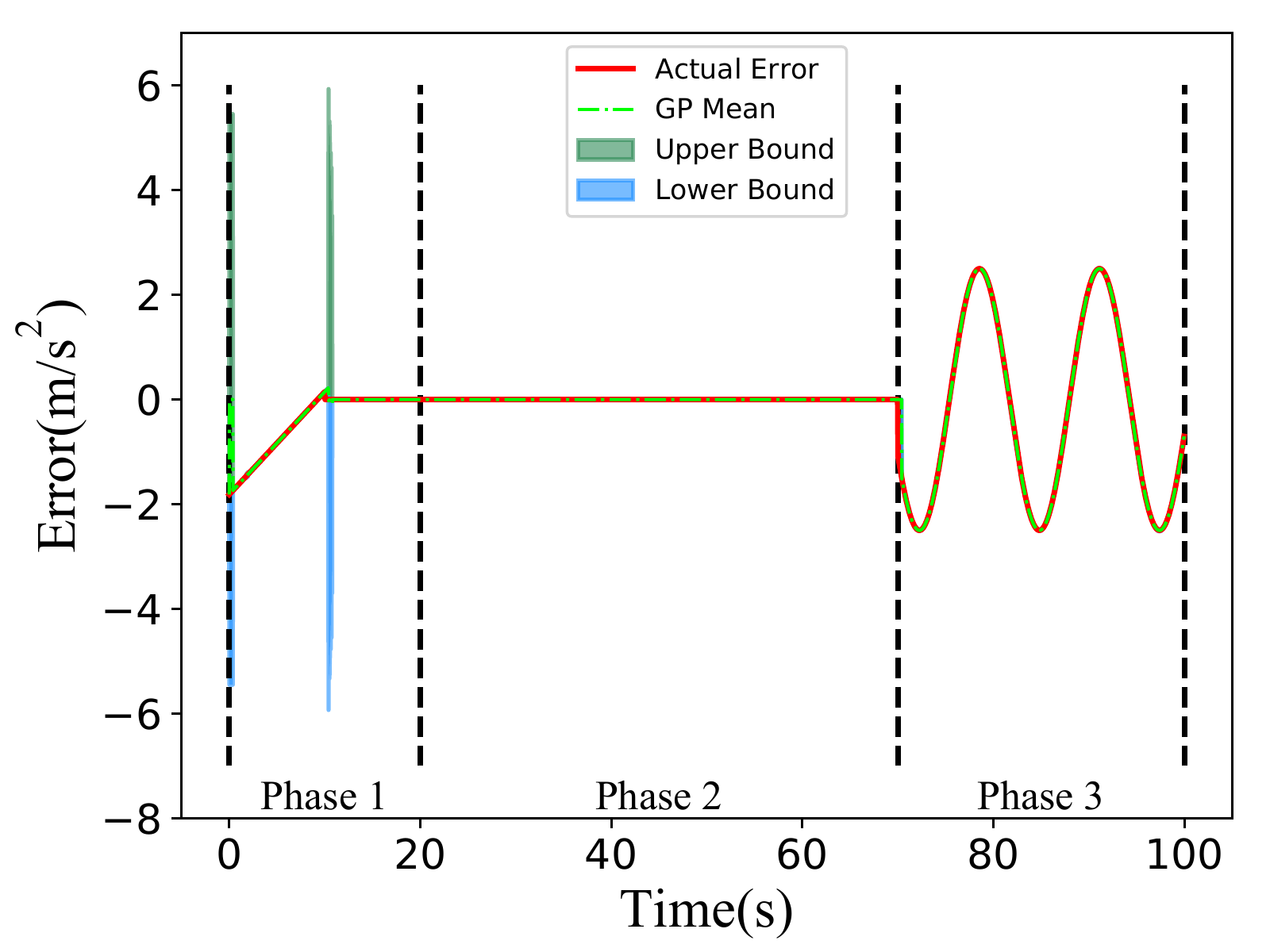}
		}\hspace{-2mm}
		\subfigure[]{
			\label{fig:err2}
			\includegraphics[scale=0.265]{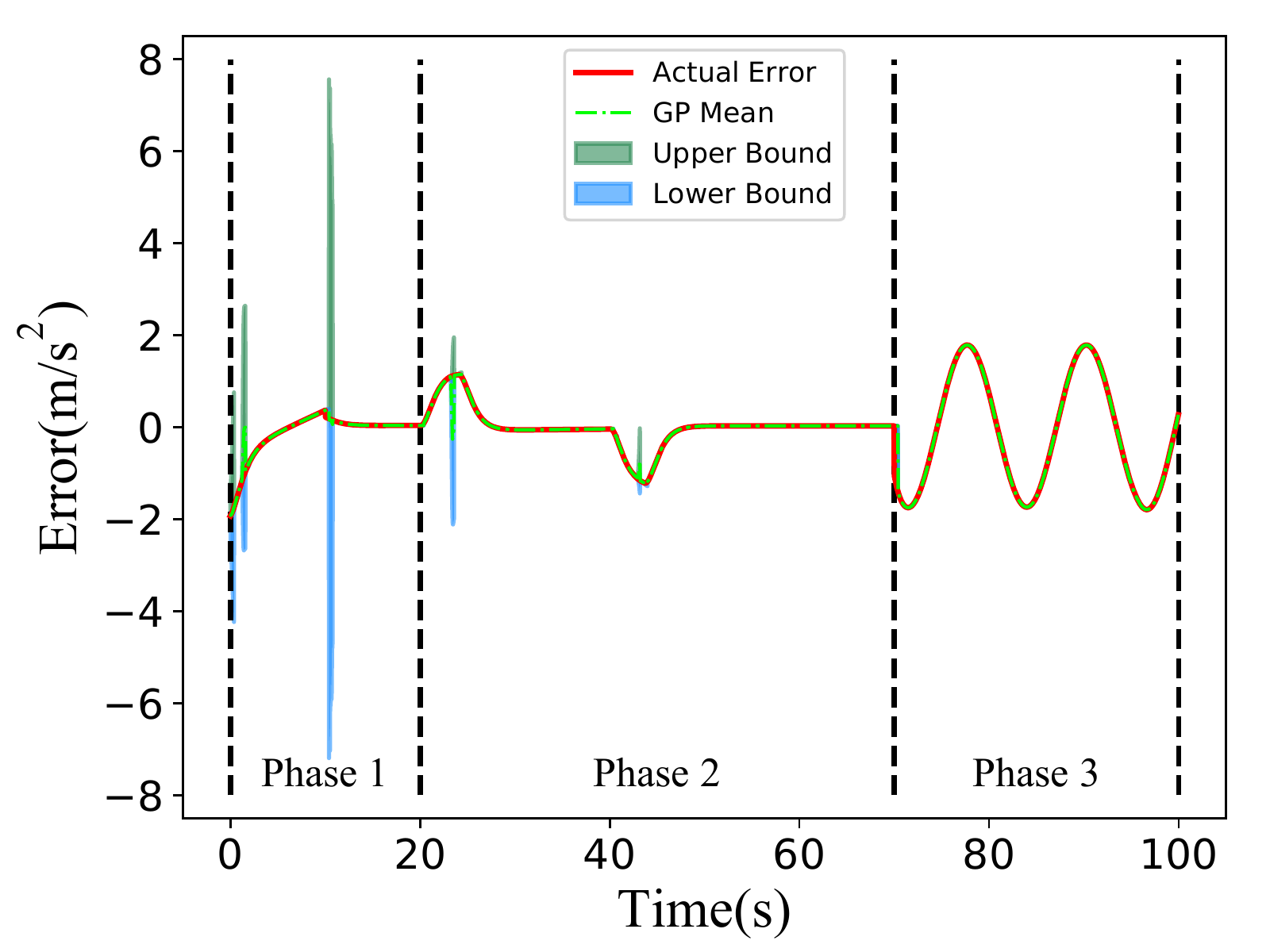}
		}\hspace{-4mm}\vspace{-3mm}

		\caption{
			The model errors of the accelerations estimated via GPs. The model errors of the acceleration of (a) the autonomous car 4, (b) the human-driven car 3.}
		\vspace{-5mm}
		\label{fig:GP}
	\end{figure}	

	\begin{figure*}[t]		
		\centering
		\hspace{-6mm}
		\subfigure[]{
			\label{fig:A1}
			\includegraphics[scale=0.26]{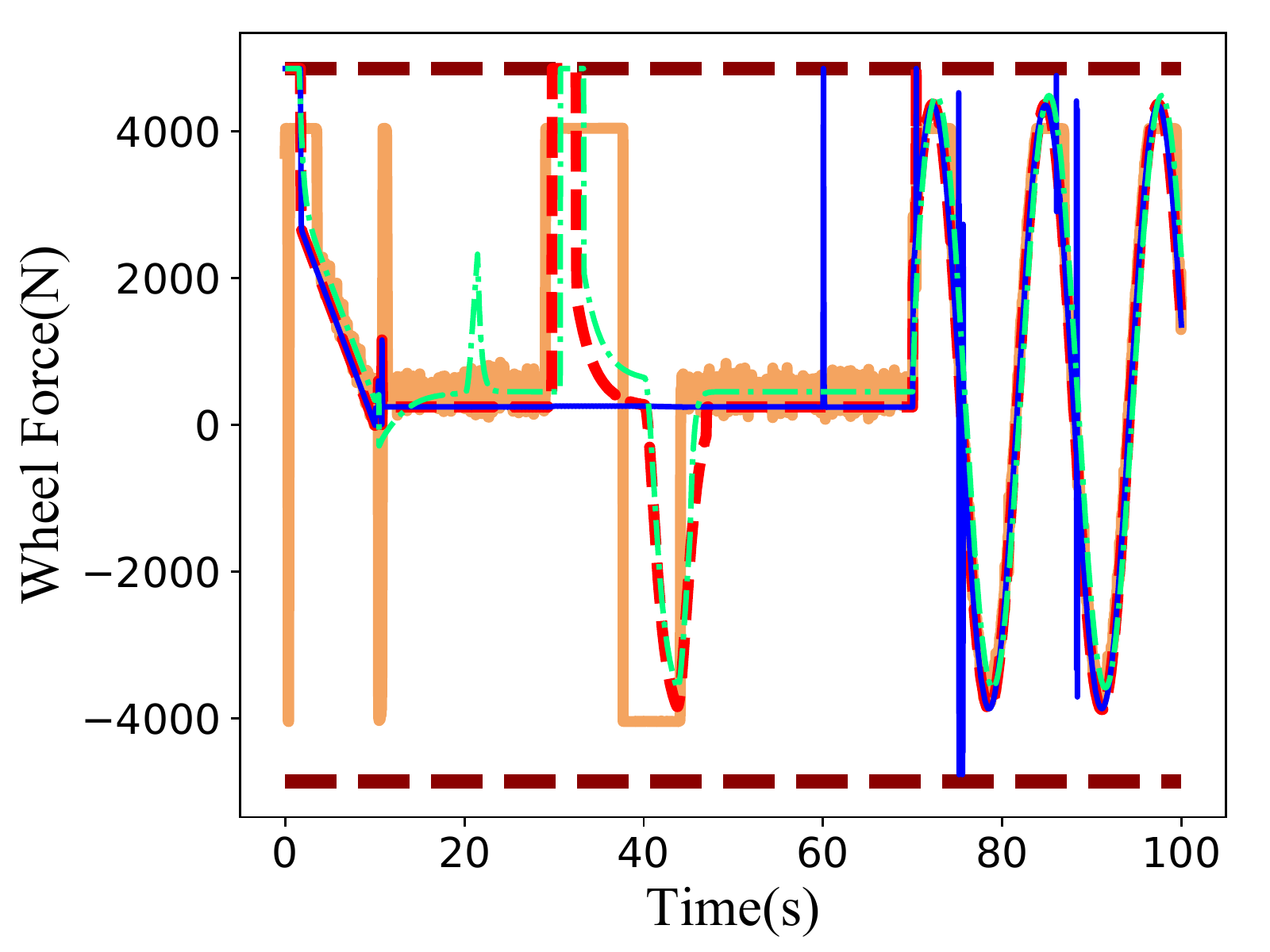}
		}\hspace{-3mm}
		\subfigure[]{
			\label{fig:Vel1}
			\includegraphics[scale=0.26]{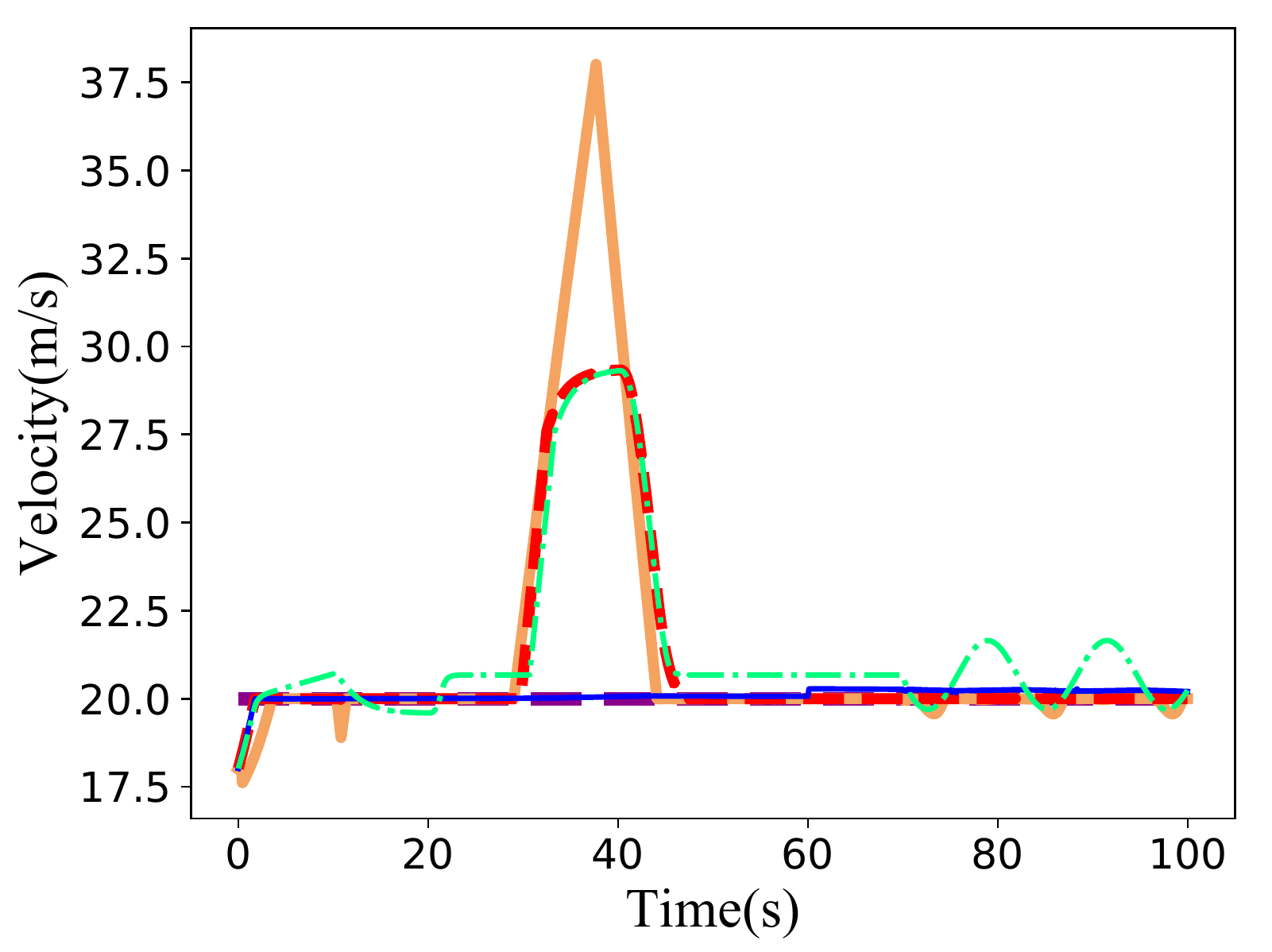}
		}\hspace{-3mm}
		\subfigure[]{
			\label{fig:DB34_1}
			\includegraphics[scale=0.26]{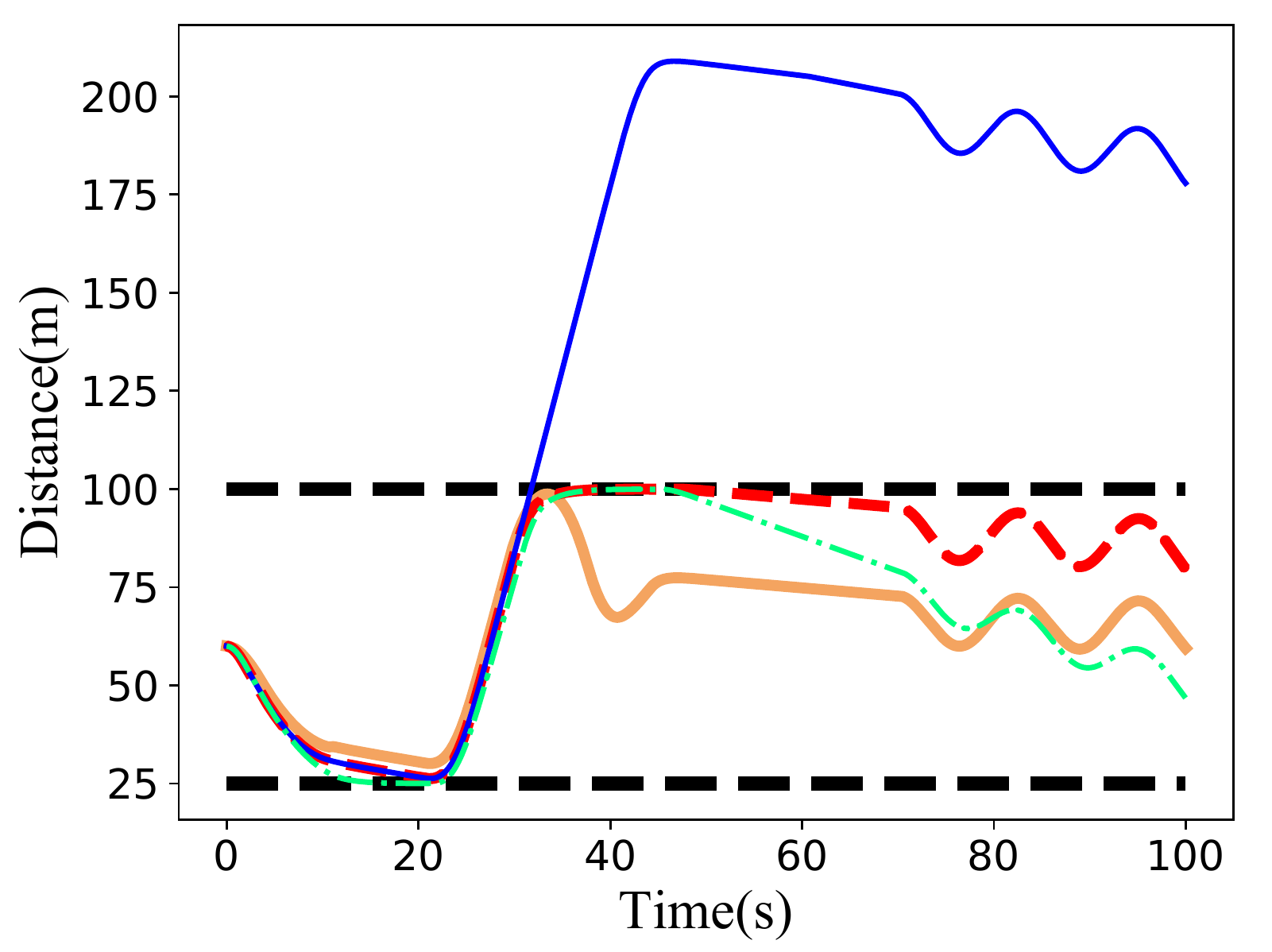}
		}\hspace{-3mm}
		\subfigure[]{
			\label{fig:TrackingError1}
			\includegraphics[scale=0.26]{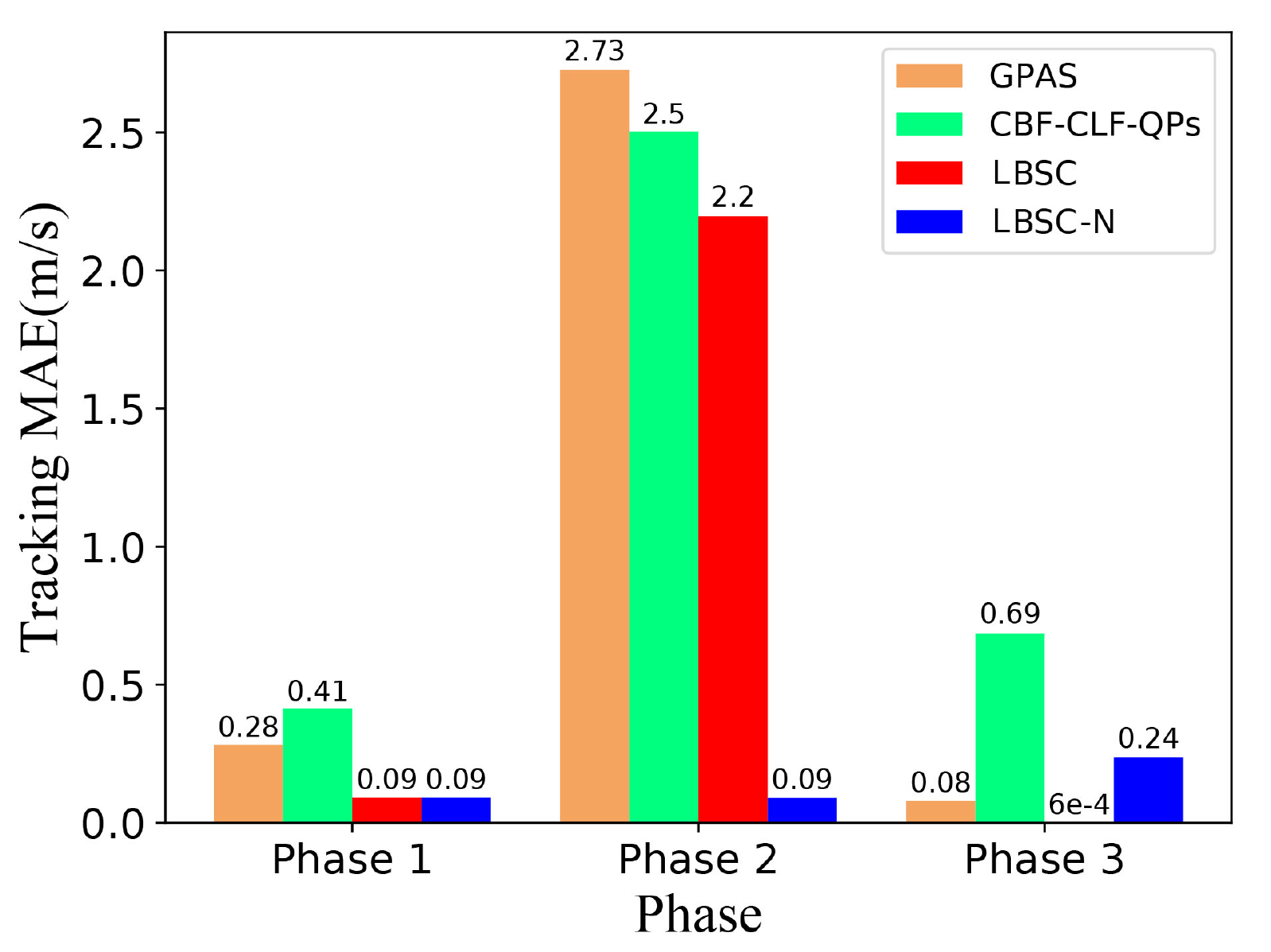}
		}\hspace{-1mm}\vspace{-2mm}
		\subfigure{
			\includegraphics[scale=0.16]{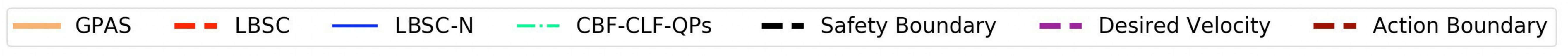}
		}		 \vspace{-3mm}
		\caption{Simulation results on the CCC system. The goal is to drive the autonomous car 4 to the desired cruising velocity while enforcing the safety constraints that the car 4 keeps a space headway ranging from 25$ m $ to 100 $m$ with its front car 3. The control input (wheel force) and velocity of the autonomous car 4 are shown in~\ref{fig:A1} and~\ref{fig:Vel1} , respectively. The space headway between car 4 and car 3 is shown in~\ref{fig:DB34_1}. The tracking MAE between the desired and tracked velocity of the car 4 is shown in~\ref{fig:TrackingError1}.}
		 \vspace{-6.5mm}
		\label{fig:CASE synthesize}
	\end{figure*} 
	\subsection{Quantitative Experiments}
	We build a CCC simulation environment by Python 3.6 to numerically validate the performance of the proposed LBSC controller. The python library CVXOPT is utilized to solve the QP problem in (\ref{LBSC_CCC}), and the average time of solving this QP problem is 2.45 $ms$. 
	In the experiments, the time of an episode is set as 100 $s$, and the control frequency is 50 $Hz$. The average time-consumption of the proposed LBSC algorithm is 49 $s$ for each episode and each control step takes 9.8$ms$ on average, which enables the LBSC algorithm to provide real-time control for the autonomous car 4.
	
	In the simulations, the car 4 is controlled to cruise at the desired speed of $v_{d} = 20 m/s$ while keeping a safe space headway ranging from 25$m$ to 100 $m$ within its front car 3. The performance of the various algorithms is studied and compared as shown in Fig.~\ref{fig:CASE synthesize}. The MAE defined in (\ref{MAE}) is used to evaluate the tracking errors in each phase.

	As shown in Fig.~\ref{fig:GP}, the actual model errors of accelerations of the car 3 and the car 4 both lie within the high confidence uncertainty bounds estimated by GPs in each phase, indicating that GPs can model the unknown discrepancies between the prior model and the actual system. The estimated uncertainties increase sharply at the beginning of simulations and $t = 10s$ when the friction coefficient changes, and decrease quickly with more data gathered. It shows that GPs can learn from experiences and provide online estimations. These estimated errors and uncertainty bounds are incorporated into the LBSC controller to help it adapts to new environment changes and generate smooth controls as depicted in Figs.~\ref{fig:A1}.

	It can be seen from Fig.~\ref{fig:Vel1}, in $\textbf{phase 1}$, the proposed LBSC approach enables the autonomous car 4 to accelerate to its desired cruising speed of 20 $m/s$ and illustrates faster convergence performance than the GPAS method~\cite{Teck2018GaussianPA}.
	Meanwhile, as shown in Fig.~\ref{fig:DB34_1}, the distance of the car 4 to its front car 3 is kept within the safe range from 25 $m$ to 100 $m$.
	
	In $\textbf{phase 2}$, the lead car 1 urgently accelerate then decelerate. In such situations, as illustrated in Fig.~\ref{fig:DB34_1}, the LBSC method strictly ensures the specified space headway constraints. On the other hand, as shown in Fig.~\ref{fig:TrackingError1}, the tracking MAE of the LBSC controller in $\textbf{phase 2}$ is much larger than that in $\textbf{phase 1}$. It indicates that the LBSC approach enables the safety to be a higher priority when there is a conflict between safety and tracking stability. 
	
	In $\textbf{phase 3}$, the velocity of the lead car 1 continuously varies. As shown in Fig.~\ref{fig:CASE synthesize}, the tracking stability and safety of the car 4 are both guaranteed by the LBSC method. On the contrary, it shows from Fig.~\ref{fig:DB34_1} that the LBSC-N algorithm violates the safety constraints in $\textbf{phase 2}$ and $\textbf{phase 3}$, although its tracking errors are smaller compared to other methods in $\textbf{phase 2}$ as shown in Fig.~\ref{fig:TrackingError1}. 
	
	Furthermore, it can be shown from Figs.~\ref{fig:A1} and~\ref{fig:Vel1} that the LBSC method generates smoother control input (wheel force) and velocity curves than those of the GPAS method. Besides, as shown in Fig.~\ref{fig:TrackingError1}, the tracking MAE of the LBSC method is much less than the GPAS and CBF-CLF-QPs methods~\cite{Ames2017ControlBF}. It implies that the LBSC controller can achieve better tracking performance.

	\section{Conclusion}
	\label{section:conclusion}	
	
	 In this paper, a learning-based control algorithm (LBSC) was proposed for nonlinear safety-critical systems subject to control input constraints under model uncertainties. GPs were employed to learn the model errors online between the nominal model and the actual system dynamics. Specifically, using the uncertainty bound estimated via GPs, a safety barrier and a stability constraint were proposed to achieve safety (represented by ZCBF) guarantees and stabilization objective (represented by ES-CLF) for the uncertain safety-critical systems, respectively. Considering control constraints, the safety barrier and stability constraint were unified in a QP to mediate the tradeoff between safety guarantees and tracking performance. The effectiveness of the LBSC algorithm was illustrated on the safety-critical CCC system under the high change rate of the acceleration and model uncertainties. In future work, the experimental validations of the proposed LBSC method will be performed on wheeled and aerial robots.

	

	\bibliographystyle{IEEEtran}
	\bibliography{egbib}

\begin{thebibliography}{10}
\providecommand{\url}[1]{#1}
\csname url@samestyle\endcsname
\providecommand{\newblock}{\relax}
\providecommand{\bibinfo}[2]{#2}
\providecommand{\BIBentrySTDinterwordspacing}{\spaceskip=0pt\relax}
\providecommand{\BIBentryALTinterwordstretchfactor}{4}
\providecommand{\BIBentryALTinterwordspacing}{\spaceskip=\fontdimen2\font plus
\BIBentryALTinterwordstretchfactor\fontdimen3\font minus
  \fontdimen4\font\relax}
\providecommand{\BIBforeignlanguage}[2]{{%
\expandafter\ifx\csname l@#1\endcsname\relax
\typeout{** WARNING: IEEEtran.bst: No hyphenation pattern has been}%
\typeout{** loaded for the language `#1'. Using the pattern for}%
\typeout{** the default language instead.}%
\else
\language=\csname l@#1\endcsname
\fi
#2}}
\providecommand{\BIBdecl}{\relax}
\BIBdecl

\bibitem{hovakimyan20111}
N.~Hovakimyan, C.~Cao, E.~Kharisov, E.~Xargay, and I.~M. Gregory, ``L 1
  adaptive control for safety-critical systems,'' \emph{IEEE Control Systems
  Magazine}, vol.~31, no.~5, pp. 54--104, 2011.

\bibitem{Knight2002SafetyCS}
J.~C. Knight, ``Safety critical systems: challenges and directions,''
  \emph{Proceedings of the 24th International Conference on Software
  Engineering. ICSE 2002}, pp. 547--550, 2002.

\bibitem{Teck2018GaussianPA}
T.~Y. Teck, A.~Kunapareddy, and M.~Kobilarov, ``Gaussian process adaptive
  sampling using the cross-entropy method for environmental sensing and
  monitoring,'' \emph{2018 IEEE International Conference on Robotics and
  Automation (ICRA)}, pp. 6220--6227, 2018.

\bibitem{Ostafew2016RobustCL}
C.~J. Ostafew, A.~P. Schoellig, and T.~D. Barfoot, ``Robust constrained
  learning-based nmpc enabling reliable mobile robot path tracking,'' \emph{I.
  J. Robotics Res.}, vol.~35, pp. 1547--1563, 2016.

\bibitem{schaal2010learning}
S.~Schaal and C.~G. Atkeson, ``Learning control in robotics,'' \emph{IEEE
  Robotics \& Automation Magazine}, vol.~17, no.~2, pp. 20--29, 2010.

\bibitem{Ames2019ControlBF}
A.~D. Ames, S.~Coogan, M.~Egerstedt, G.~Notomista, K.~Sreenath, and P.~Tabuada,
  ``Control barrier functions: Theory and applications,'' \emph{2019 18th
  European Control Conference (ECC)}, pp. 3420--3431, 2019.

\bibitem{Ames2017ControlBF}
A.~D. Ames, X.~Xu, J.~W. Grizzle, and P.~Tabuada, ``Control barrier function
  based quadratic programs for safety critical systems,'' \emph{IEEE
  Transactions on Automatic Control}, vol.~62, pp. 3861--3876, 2017.

\bibitem{Xu2015RobustnessOC}
X.~Xu, P.~Tabuada, J.~W. Grizzle, and A.~D. Ames, ``Robustness of control
  barrier functions for safety critical control,'' in \emph{ADHS}, 2015.

\bibitem{Agrawal2017DiscreteCB}
A.~Agrawal and K.~Sreenath, ``Discrete control barrier functions for
  safety-critical control of discrete systems with application to bipedal robot
  navigation,'' in \emph{Robotics: Science and Systems}, 2017.

\bibitem{wu2016safety}
G.~Wu and K.~Sreenath, ``Safety-critical control of a 3d quadrotor with
  range-limited sensing,'' in \emph{Dynamic Systems and Control Conference},
  vol. 50695.\hskip 1em plus 0.5em minus 0.4em\relax American Society of
  Mechanical Engineers, 2016, p. V001T05A006.

\bibitem{Ames2014ControlBF}
A.~D. Ames, J.~W. Grizzle, and P.~Tabuada, ``Control barrier function based
  quadratic programs with application to adaptive cruise control,'' \emph{53rd
  IEEE Conference on Decision and Control}, pp. 6271--6278, 2014.

\bibitem{khalil2002nonlinear}
H.~K. Khalil, ``Nonlinear systems,'' \emph{Upper Saddle River}, 2002.

\bibitem{Wang2017SafeLO}
L.~Wang, E.~Theodorou, and M.~Egerstedt, ``Safe learning of quadrotor dynamics
  using barrier certificates,'' \emph{2018 IEEE International Conference on
  Robotics and Automation (ICRA)}, pp. 2460--2465, 2017.

\bibitem{Rasmussen2005GaussianPF}
C.~E. Rasmussen and C.~K.~I. Williams, ``Gaussian processes for machine
  learning,'' in \emph{Adaptive computation and machine learning}, 2005.

\bibitem{Berkenkamp2016SafeLO}
F.~Berkenkamp, R.~Moriconi, A.~P. Schoellig, and A.~Krause, ``Safe learning of
  regions of attraction for uncertain, nonlinear systems with gaussian
  processes,'' \emph{2016 IEEE 55th Conference on Decision and Control (CDC)},
  pp. 4661--4666, 2016.

\bibitem{scholkopf2002learning}
B.~Sch{\"o}lkopf, A.~J. Smola, F.~Bach \emph{et~al.}, \emph{Learning with
  kernels: support vector machines, regularization, optimization, and
  beyond}.\hskip 1em plus 0.5em minus 0.4em\relax MIT press, 2002.

\bibitem{srinivas2012information}
N.~Srinivas, A.~Krause, S.~M. Kakade, and M.~W. Seeger, ``Information-theoretic
  regret bounds for gaussian process optimization in the bandit setting,''
  \emph{IEEE Transactions on Information Theory}, vol.~58, no.~5, pp.
  3250--3265, 2012.

\bibitem{AmesRapidly}
A.~D. Ames, K.~Galloway, K.~Sreenath, and J.~W. Grizzle, ``Rapidly
  exponentially stabilizing control lyapunov functions and hybrid zero
  dynamics,'' \emph{IEEE Transactions on Automatic Control}, vol.~59, no.~4,
  pp. 876--891.

\bibitem{He2018DatabasedFO}
C.~R. He, J.~I. Ge, and G.~Orosz, ``Data-based fuel-economy optimization of
  connected automated trucks in traffic,'' \emph{2018 Annual American Control
  Conference (ACC)}, pp. 5576--5581, 2018.

\bibitem{Pedregosa2011ScikitlearnML}
F.~Pedregosa, G.~Varoquaux, A.~Gramfort, V.~Michel, B.~Thirion, O.~Grisel,
  M.~Blondel, G.~Louppe, P.~Prettenhofer, R.~Weiss, V.~Dubourg, J.~VanderPlas,
  A.~Passos, D.~Cournapeau, M.~Brucher, M.~Perrot, and E.~Duchesnay,
  ``Scikit-learn: Machine learning in python,'' \emph{J. Mach. Learn. Res.},
  vol.~12, pp. 2825--2830, 2011.

\end{thebibliography}
	\printindex
\end{document}